%% file: main.tex
\documentclass[letterpaper, 10 pt, conference]{ieeeconf}  % Comment this line out if you need a4paper

\IEEEoverridecommandlockouts                              % This command is only needed if 
\usepackage{xcolor}
\usepackage{multirow}
\usepackage{gensymb}
\usepackage{booktabs} 
\usepackage{cite}
\usepackage{algorithm}
\usepackage{algorithmic}
\usepackage{makecell}
\usepackage{textcomp}
\usepackage{graphics} % for pdf, bitmapped graphics files
\usepackage{epsfig} % for postscript graphics files
\usepackage{times} % assumes new font selection scheme installed
\usepackage{amsmath} % assumes amsmath package installed
\usepackage{amssymb}  % assumes amsmath package installed
\usepackage{xspace}
\usepackage[breaklinks=true,bookmarks=false]{hyperref}

\definecolor{outlier}{rgb}{1, 0.55, 0} 
\definecolor{lidar}{rgb}{0, 0.75, 1} 
\definecolor{radar}{rgb}{1,0,0.75}

\newcommand{\sysname}{\texttt{RaTrack}\xspace}

\title{\LARGE \bf
RaTrack: Moving Object Detection and Tracking with \\ 4D Radar Point Cloud
}

\author{Zhijun Pan\authorrefmark{2}, Fangqiang Ding\authorrefmark{2}, Hantao Zhong\authorrefmark{2}, and Chris Xiaoxuan Lu\authorrefmark{1}
\thanks{\authorrefmark{2}Equal Contribution, listed randomly. \authorrefmark{1}Corresponding author.}
\thanks{Zhijun Pan is with the Computer Science Research Centre, Royal College of Art, United Kingdom. This work was partly done when he was with the School of Informatics, University of Edinburgh.}
\thanks{Fangqiang Ding is with the School of Informatics, University of Edinburgh, United Kingdom. This research is supported by the EPSRC, as part
of the CDT in Robotics and Autonomous Systems hosted at the Edinburgh Centre of Robotics (EP/S023208/1).}
\thanks{Hantao Zhong is with the Department of Computer Science and Technology, University of Cambridge, United Kingdom.}
\thanks{Chris Xiaoxuan Lu is with the Department of Computer Science, University College London, United Kingdom.}
}

\begin{document}

\maketitle
\thispagestyle{empty}
\pagestyle{empty}

%%%%%% ABSTRACT
%%%%%%%%%%%%%%%%%%%%%%%%%%%%%%%%%%%%%%%%%%%%%%%%%%%%%%%%%%%%%%%%%%%%%%%%%%%%%%%%
\begin{abstract}
Mobile autonomy relies on the precise perception of dynamic environments. Robustly tracking moving objects in 3D world thus plays a pivotal role for applications like trajectory prediction, obstacle avoidance, and path planning. While most current methods utilize LiDARs or cameras for Multiple Object Tracking (MOT), the capabilities of 4D imaging radars remain largely unexplored. Recognizing the challenges posed by radar noise and point sparsity in 4D radar data, we introduce \sysname, an innovative solution tailored for radar-based tracking. Bypassing the typical reliance on specific object types and 3D bounding boxes, our method focuses on motion segmentation and clustering, enriched by a motion estimation module. Evaluated on the View-of-Delft dataset, \sysname showcases superior tracking precision of moving objects, largely surpassing the performance of the state of the art. We release our code and model at \href{https://github.com/LJacksonPan/RaTrack}{https://github.com/LJacksonPan/RaTrack}.
\end{abstract}

%%%%%% BODY TEXT
%%%%%%%%%%%%%%%%%%%%%%%%%%%%%%%%%%%%%%%%%%%%%%%%%%%%%%%%%%%%%%%%%%%%%%%%%%%%%%%%

\input{section/introduction}
\input{section/related_work}

\input{section/problem_formulation}
\input{section/method}

\input{section/experiment}

\input{section/conclusion}

%%%%%%%%%%%%%%%%%%%%%%%%%%%%%%%%%%%%%%%%%%%%%%%%%%%%%%%%%%%%%%%%%%%%%%%%%%%%%%%%
\vspace{1em}
\bibliographystyle{IEEEtran}
\bibliography{reference}

\end{document}

%% file: section/introduction.tex
\section{Introduction}

Ensuring accurate perception of dynamic environments is pivotal for mobile autonomy. A crucial task in this domain is the consistent and robust tracking of moving objects in 3D space for autonomous vehicles. This ability acts as a cornerstone for subsequent autonomy tasks such as trajectory prediction~\cite{zhao2021tnt,deo2022multimodal,wei2020energy}, obstacle avoidance~\cite{li2021prediction,lin2020robust,zhang2019real}, and path planning~\cite{petres2007path,yonetani2021path,inotsume2020robust}.

State-of-the-art techniques in moving object tracking or Multiple Object Tracking (MOT) primarily use on-vehicle LiDARs~\cite{weng20203d, yin2021center, chiu2021probabilistic}, cameras~\cite{cc3dt,hu2022monocular,ding2020automatic}, or their fusion~\cite{shenoi2020jrmot, kim2021eagermot, zeng2021cross}. Surprisingly, the potential of 4D mmWave radars remains under-explored. As an emerging automotive sensor, 4D mmWave radar is gaining traction due to its improved imaging ability, resilience against challenging weather and illumination conditions (e.g., fog, dust, darkness, glare), ability to measure object velocities and cost-effectiveness. These merits make 4D radar an appealing and robust supplement, or even alternative, to automotive LiDARs.

However, integrating 4D radars into moving object tracking presents non-trivial challenges. The prevalent approaches~\cite{weng20203d, yin2021center, chiu2021probabilistic, liang2020pnpnet, kim2021eagermot} often follow the \emph{tracking-by-detection} paradigm. Such a paradigm involves first detecting objects in each frame independently and then linking these \emph{detected object types and 3D bounding boxes} across consecutive frames to form continuous object trajectories. Key to the \emph{tracking-by-detection} success depends on the detection accuracy. 
This paradigm struggles when adapted to 4D radar data, due to the inherent radar noise and point sparsity, undermining accurate type classification and bounding box regression. Specifically, the non-negligible multi-path noises in radar data complicate the correct identification of objects while the sparsity of radar point clouds makes the object property (e.g. extension and orientation) regression even more difficult. As exhibited in~\cite{palffy2022multi}, the mAP performance~\cite{geiger2013vision} of the 4D radar detection method is only 38.0, $\sim$40\% inferior to its LiDAR counterpart in the same scene. Such poor 3D detection performance compromises the reliability of 4D radar-based tracking in real-world scenarios. 

To address this, we present \sysname, a first-of-its-kind tailored solution for moving object tracking using 4D automotive radars. Our approach stems from a critical insight: \emph{for effective multi-object tracking, class-agnostic detection is often adequate, and the conventional reliance on 3D bounding boxes becomes redundant if distinct point clusters can be utilized}. Driven by this understanding, we restructure the moving object detection challenge into simpler motion segmentation and clustering tasks. This restructuring allows us to sidestep the complex tasks of object type classification and bounding box regression typically encountered with 4D radars.
To further enhance our method's performance, we integrate a point-wise motion estimation module, which enriches the inherently sparse radar data with point-level scene flow. Building on this, our data association module is precisely adapted to our clustering method and is calibrated to weigh different features for optimal matching. Our solution is architected as an end-to-end trainable network, with its training modelled as a multi-task learning endeavour. This encompasses motion segmentation, scene flow estimation, and affinity matrix computation.

Extensive experiments on the View-of-Delft dataset~\cite{palffy2022multi} validate the superiority of \sysname over existing techniques in moving object detection and tracking precision. Moreover, our results underscore the merits of the cluster-based detection method and the instrumental role of scene flow estimation in both detection and data association phases. 

%% file: section/related_work.tex
\section{Related Works}
Given the absence of prior work on 4D radar-based moving object tracking, we will touch on existing research in general 3D MOT. As an uplift of 2D MOT~\cite{milan2016mot16,ciaparrone2020deep,wang2020towards,voigtlaender2019mots,meinhardt2022trackformer} in the 3D space, 3D MOT has attracted increasing interests due to its significant application to autonomous systems. Most online 3D MOT solutions adopt a tracking-by-detection approach, focusing on 3D bounding box detection and data association. The core of data association lies in extracting comprehensive tracking cues and matching new detections with previous tracklets.

\noindent\textbf{3D bounding box detection.} The premise of applying the tracking-by-detection pipeline is accurate 3D bounding box estimation. Thanks to recent advances in 3D object detection, many off-the-shelf detectors~\cite{ku2018joint,zhu2019class,shi2019pointrcnn,Yang_2018_CVPR,yin2021center,shi2020point} have already been employed as the front-end of modern tracking systems. According to their input modalities used for 3D detection, current 3D MOT systems can be classfied into LiDAR point cloud-based~\cite{weng20203d,baser2019fantrack,poschmann2020factor,weng2020gnn3dmot,benbarka2021score,chiu2021probabilistic,pang2021simpletrack,zaech2022learnable,wen2022pf,kim2022polarmot,sadjadpour2022shasta}, image-based~\cite{cc3dt,hu2022monocular,chaabane2021deft,zhou2020tracking} and LiDAR-image fusion-based~\cite{shenoi2020jrmot,liang2020pnpnet,kim2021eagermot,zeng2021cross} methods. 
%Note that their detection step can be replaced by any 3D detectors if needed because of their independence from the tracking process. 
Different from these approaches, our method takes only 4D radar point clouds as the input and detects object instances as clusters of points instead of 3D bounding boxes for tracking.

\noindent\textbf{Tracking cues.} %Different types of cues can be exploited from the detection results or input data for data association. 
To exploit the 3D motion information, AB3DMOT~\cite{weng20203d} proposes a baseline method that models the motion of objects with a Kalman filter and predicts the displacement with a constant velocity model. The same strategy and its variants~\cite{chiu2021probabilistic} are followed by later works~\cite{shenoi2020jrmot,pang2021simpletrack,kim2021eagermot,benbarka2021score,zaech2022learnable} to induce motion cues for association. Another common strategy is to directly regress object velocities from the detectors~\cite{yin2021center,zeng2021cross,wen2022pf}. In~\cite{weng2020gnn3dmot,liang2020pnpnet}, latent motion features are extracted by an LSTM network for tracked objects. Apart from motion cues, object appearance features are usually learned from neural networks, either from images~\cite{baser2019fantrack}, LiDAR point clouds~\cite{liang2020pnpnet,wen2022pf,sadjadpour2022shasta} or both of them~\cite{shenoi2020jrmot,weng2020gnn3dmot,zeng2021cross,chiu2021probabilistic}, for data association. Unlike prior works, we estimate per-point scene flow vectors to obtain motion cues, which can not only help to match objects with similar motion but also benefit moving object detection. %More specifically, scene flow can also be applied as an attached feature for clustering points in our tracking-by-clustering pipeline.
Besides scene flow, we aggregate complementary point-level features from each cluster for robust data association.

\noindent\textbf{Track-detection matching.}  Given object motion or appearance cues, most methods generate an affinity map based on object motion or appearance cues, capturing matching scores for potential track-detection pairs. Some methods~\cite{baser2019fantrack,shenoi2020jrmot,weng20203d} use traditional distance metrics like cosine or $L_2$ distances, while others~\cite{liang2020pnpnet,weng2020gnn3dmot} employ networks for learnable distance metrics. Assignments are typically resolved using the Hungarian~\cite{kuhn1955hungarian} or greedy algorithms. Recent techniques~\cite{zaech2022learnable,kim2022polarmot} employ graph structures and Neural Message Passing~\cite{braso2020learning} for more direct associations. In our approach, inspired by~\cite{liang2020pnpnet,weng2020gnn3dmot}, we use MLP networks to estimate cluster-pair similarities. Uniquely, we adopt the differentiable Sinkhorn algorithm~\cite{sinkhorn1967concerning} for bipartite matching, rendering the data association process fully differentiable, enhancing tasks like trajectory prediction and planning.

%% file: section/problem_formulation.tex
\section{Problem Formulation}

\noindent \textbf{Scope}. In a standard 3D MOT setup, all objects of interest, such as cars and motorcycles, are tracked regardless of their motion status~\cite{weng20203d,pang2021simpletrack,kim2022polarmot}. Contrarily, this work solely focuses on the moving objects. This focus stems from the premise that dynamic entities hold greater significance for tracking than static ones. Additionally, the inherent ability of radar sensors to measure velocity makes it a trivial task to distinguish between moving and stationary objects.

\noindent \textbf{Notation}. Given this context, we consider the problem of online \emph{moving object detection and tracking} with 4D automotive radar. The input is an ordered 4D radar point cloud sequence $\mathcal{P} = \{\mathbf{P}^t\}_{t=1}^{T}$ comprised of $T$ frames captured by the same radar sensor mounted on a moving vehicle. A frame $\mathbf{P}^t = [\mathbf{p}^t_1; ...; \mathbf{p}^t_i; ...; \mathbf{p}^t_{N^t}]$ contains $N^t$ radar points. Each radar point $\mathbf{p}_i^t$ is characterised by its 3D position $\mathbf{x}_i^t\in\mathbb{R}^3$ in the metric space and auxiliary velocity features $\mathbf{v}_i^t = [v_{r,i}^t, v_{c,i}^t]$, where 
$v_{r,i}^t$ and $v_{c,i}^t$ are the measured relative radial velocity (RRV) and its ego-motion (assumed known) compensated variant. Given each radar point cloud $\mathbf{P^t}$ from the stream, our objective is to detect multiple moving objects $\mathbf{D}^{t}=\{\mathbf{d}_k^t\}_{k=1}^{K^t}$  in a class-agnostic manner without the need to regress their 3D bounding boxes. These detected objects are then associated with objects tracked in the previous frame $\mathbf{O}^{t-1}=\{\mathbf{o}_m^{t-1}\}_{m=1}^{M^t}$. The result of this process is a set of updated objects $\mathbf{O}^{t}$ in track for the current frame. 

%% file: section/method.tex
\section{Proposed Method}\label{method}

\begin{figure*}[!tbp]
  \centering
  \includegraphics[scale=0.165]{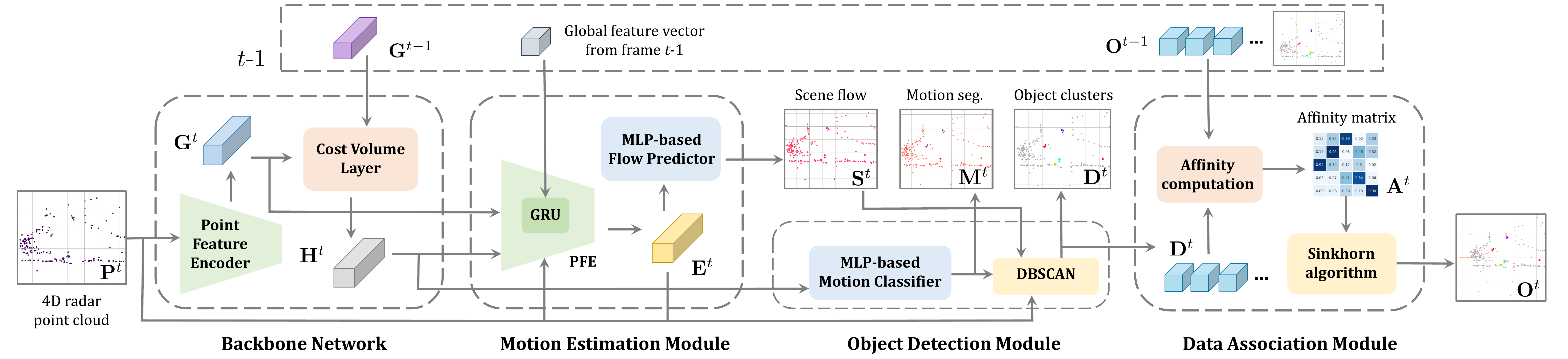}
  \vspace{-0.5em}
  \caption{Overall network pipeline of \sysname. Note that radar 3D points are shown in the bird's eye view for visualization.}
  % ivan, please see the wechat
  \label{fig:pipeline}
  \vspace{-1em}
\end{figure*}

\subsection{Overview}
We introduce \sysname, a generic learning-based framework bespoken for 4D radar-based moving object detection and tracking. As seen in Fig.~\ref{fig:pipeline}, in our network pipeline, we first apply a backbone (c.f. Sec.~\ref{backbone}) to encode intra- and inter-frame radar point cloud features. With the extracted features, our point-wise motion estimation module (c.f. Sec.~\ref{motion}) infers point-level scene flow as an explicit complement to augment the latent features of radar point clouds. Our advocated idea of class-agnostic detection without bounding boxes is introduced in the object detection module (c.f. Sec.~\ref{detection}), in which moving points are first identified and then used to detect moving objects 
via clustering. Finally, our data association module (c.f. Sec.~\ref{association}) computes the affinity matrix with a learnable distance metric and then optimises the bipartite matching problem. The entire network is end-to-end trainable with a multi-task loss that incorporates three supervised subtasks: motion segmentation, scene flow estimation, and affinity matrix computation. 

\subsection{Backbone Network}\label{backbone} On receiving a new 4D radar point cloud $\mathbf{P^t}$, our backbone neural network is used to extract representative latent features for each radar point to facilitate subsequent tasks, e.g., motion segmentation and scene flow estimation. To this end, we first extract point-level local-global features $\mathbf{G}^t$ using a point feature encoder (PFE), which comprises a) three set abstraction layers~\cite{qi2017pointnet2} to extract local features at different scales in parallel, b) three MLP-based feature propagation layer to map local features into high-level representations, and c) the max-pooling operation to aggregate the global feature vector that is attached to per-point features. To further encode inter-frame point motion for the current frame, we recall the features $\mathbf{G}^{t-1}$ from the last frame and correlate features across two frames by the cost volume layer~\cite{wu2020pointpwc}, as seen in Fig.~\ref{fig:pipeline}. The output is the cost volume $\mathbf{H}^t$ that represents the motion information for each point in $\mathbf{P}^t$.

\subsection{Motion Estimation Module}\label{motion}
Radar sensors' inherent sparsity and noise result in latent features from radar point clouds that are deficient in informative geometric cues, complicating object detection and tracking. To address this, we introduce a point-wise motion estimation module that explicitly determines per-point motion vectors. Contrary to the conventional scene flow approach, which estimates per-point motion vectors in a forward direction (from frame $t$ to $t+1$), we opt for a backward estimation (from frame $t$ to $t-1$). This not only negates tracking latency but also ensures that the estimated flow vectors correspond to the points in the current frame $t$.

Before decoding scene flow, we first aggregate mixed features by integrating diverse backbone outputs: input point features $\mathbf{P}^t$, local-global features $\mathbf{G}^t$ and cost volume $\mathbf{H}^t$. Subsequently, another PFE is employed, aiming to facilitate information exchange among these diverse features and enhance their spatial coherence, producing the flow embedding $\mathbf{E}^t$ as shown in Fig.~\ref{fig:pipeline}. Notably, within this PFE, a GRU network~\cite{cho2014properties} is utilized to introduce temporal information into the global feature vector prior to its association with per-point features. The scene flow $\mathbf{S}^t = [\mathbf{s}_1^t; ...; \mathbf{s}_i^t; ...; \mathbf{s}_{N^t}^t]\in\mathbb{R}^{N^t\times 3} $ is finally decoded with an MLP-based flow predictor. This module plays a crucial role in our framework and is leveraged to augment per-point features before clustering (c.f. Sec.~\ref{detection}) and to lead up data association as an extra motion clue (c.f. Sec.~\ref{association}). 

\subsection{Object Detection Module}\label{detection}
In the classical tracking-by-detection paradigm, objects are first detected as class-specific 3D bounding boxes based on which tracklets are built up across frames. However, the performance of such approaches inevitably relies on accurate type classification and bounding box regression, which are hard to accomplish from the sparse and noisy 4D radar data. 

% To refrain from using the error-prone 3D bounding box detected from specific object types and realize what we essentially need here is just to \textit{group scattered points into clusters} that can facilitate tracking, we advocate the concept of class-agnostic object detection without bounding boxes in our solution. In this way, the error-prone bounding box detector will be replaced by the combination of motion segmentation and clustering which are more reliable for radar point clouds.

Rather than relying on the error-prone 3D bounding box detection tailored for specific object types, we emphasize the fundamental necessity to simply group scattered points into clusters for effective tracking. Consequently, we champion a class-agnostic object detection approach that eschews bounding boxes in our solution. By adopting this methodology, the fallible bounding box detector is replaced by a more dependable combination of motion segmentation and clustering, which proves to be particularly suited for radar point clouds.
In other words, we detect objects in a bottom-up fashion, where points are first classified into moving and static (i.e., motion segmentation) and those close in the latent feature space are aggregated into object clusters. For motion segmentation, we leverage the cost volume $\mathbf{H}^t$ provided by the backbone and compute the moving possibility score $c_i^t$ for each point $\mathbf{p}^t_{i}$ through an MLP-based motion classifier.
% \chris{FQ, does the classification here mean determining `static' or `dynamic'? If so, we need to make it clear. Otherwise, it contradicts with our `class-agnostic' concept}. 
The classification results are reliable enough as both the crucial RRV measurements and the inter-frame motion information are encoded by our backbone, yielding a robust motion representation. A fixed threshold $\zeta_{mov}$ is further used to separate the moving targets from the static background, resulting in a motion segmentation mask $\mathbf{M}^t = \{m^t_i\in\{0,1\}\}_{i=1}^{N_t}$ as exhibited in Fig.~\ref{fig:pipeline}. To delineate the boundaries of moving objects from the pinpointed moving points, we employ the classic clustering algorithm, DBSCAN~\cite{ester1996density}. This groups analogous points into object clusters, represented as ${\mathbf{D}}^t = \{{\mathbf{d}}_k^t\}_{k=1}^{{K}^t}$ and we term $\mathbf{D}^t$ as the detected (moving) objects hereafter. For robust clustering, we utilize the point cloud $\mathbf{P}_t$, their estimated scene flow $\mathbf{S}^t$ and the flow embedding $\mathbf{E}^t$ as the salient features and identify neighbour points. In this way, each detected object ${\mathbf{d}}_k^t$ is represented as a cluster containing a subset of points $\{\mathbf{p}^t_{k,j}\}_j$ with their corresponding scene flow and flow embedding vectors $\{[\mathbf{s}^t_{k,j}, \mathbf{e}^t_{k,j}]\}_j$. It is worth noting that we forego estimating explicit object categories (e.g., car, pedestrian). For the purposes of object tracking, such categorization is \textit{not} imperative. Instead, we identify them simply as \textit{class-agnostic} entities.

\subsection{Data Association Module}\label{association}

Given the objects \( \mathbf{D}^t \) detected in the current frame, our data association module endeavors to align them with the previously tracked objects \( \mathbf{O}^{t-1} \) from frame \( t-1 \). For end-to-end learning and inspired by \cite{liang2020pnpnet,weng2020gnn3dmot,chiu2021probabilistic}, we opt for the MLP over traditional hand-crafted metrics to derive the affinity matrix $\mathbf{A}^t\in\mathbb{R}^{K^t\times M^t}$ for bipartite matching. Such a learnable distance metric can automatically adjust the weights of different features when calculating the similarity scores. For each pair of clusters $\{\mathbf{d}^t_k, \mathbf{o}^{t-1}_m\}$, its corresponding similarity score $a^t_{k,m}$ can be computed as follows:
\begin{equation}\small
     a^t_{k,m} = {MLP}(\mathbf{l}^t_k -\mathbf{l}^{t-1}_m) 
\end{equation}
where $\mathbf{l}^t_k$, $\mathbf{l}^{t-1}_m$ are the aggregated features of two clusters respectively. Taking the object $\mathbf{d}^t_k$ as the example, to generate its aggregated features $\mathbf{l}^t_k$, we concatenate a) the average and variance of its associated point subset $\{\mathbf{p}^t_{k,j}\}_j$, b) the max-pooling of the point-level scene flow and embedding vectors $\{[\mathbf{s}^t_{k,j}, \mathbf{e}^t_{k,j}]\}_j$. This process can effectively aggregate the information for each cluster and ensure the dimension consistency given clusters with various numbers of points.

Once the affinity matrix \( \mathbf{A}^t \) is computed, we identify the optimal matching pairs based on their similarity scores. To address this optimization challenge, we employ the Sinkhorn algorithm~\cite{sinkhorn1967concerning}, as shown in Fig.~\ref{fig:pipeline}. This method involves iterative normalization of $\exp(\mathbf{A}^t)$ across both rows and columns, ensuring the entire data association process remains differentiable. Post optimization, we reassign object IDs to the successfully matched pairs, allocate new IDs for newly detected objects, and remove IDs associated with previously tracked objects absent in the current frame. Notably, we adopt the finalized scores of matched pairs as confidence scores for currently detected objects, given the inability of our detection module to provide such scores. Such confidence scores are used for certain metrics, such as AMOTA and AMOTP, which integrate results over all recall values.

\subsection{End-to-End Training}
We leverage labelled samples to end-to-end train our network with a multi-task loss of scene flow estimation, motion segmentation and affinity matrix computation:
\begin{equation}\label{loss:overall}\small
    \mathcal{L} = \alpha_1\mathcal{L}_{flow} + \alpha_2\mathcal{L}_{seg} + \alpha_3\mathcal{L}_{aff}
\end{equation}
where $\alpha_1,\alpha_2,\alpha_3$ are hyperparameters to weigh different loss functions. For scene flow loss, we compute the $L_2$ distance between the estimated scene flow $\mathbf{S}^t$ and the ground truth one $\tilde{\mathbf{S}}^t$ = $\{\tilde{\mathbf{s}}_i^t\}_{i=1}^{N^t}$:
\begin{equation}\small
    \mathcal{L}_{flow} = \frac{1}{N^t}\sum_{i}||\mathbf{s}_i^t-\tilde{\mathbf{s}}_{i}^t||_2^2
\end{equation}
Given the ground truth motion segmentation mask $\tilde{\mathbf{M}}^{t}=\{\tilde{{m}}^t_i\}_{i=1}^{N^t}$, we separately supervise the classification scores of real moving and static points using the cross-entropy to address the low ratio ($<10\%$) of moving points in point clouds. The motion segmentation loss can be written as:
\begin{equation}\label{loss:seg}\small
    \mathcal{L}_{seg} = \beta\frac{\sum_i (1-\tilde{{m}}^t_{i})\mathrm{log}(1-{c}^t_{i})}{\sum_i 1-\tilde{{m}}^t_{i}} + (1-\beta)\frac{\sum_i\tilde{{m}}^t_{i}\mathrm{log}({c}^t_{i})}{\sum_i{\tilde{m}}^t_{i}}
\end{equation}
where $\beta$ is used to balance the influence of moving and static points. To supervise the computation of the affinity matrix $\mathbf{A}^t$, we formulate the prediction of the similarity score as a binary classification (matched or unmatched) problem and compute the binary cross-entropy loss as:
\begin{equation}\footnotesize
    \mathcal{L}_{aff} = \frac{1}{K^tM^t}\sum_k 
    \sum_m \tilde{a}_{k,m}^t\mathrm{log}({a}_{k,m}^t) + (1-\tilde{a}_{k,m}^t)\mathrm{log}(1-{a}_{k,m}^t)
\end{equation}
where $\tilde{a}_{k,m}^t\in\{0,1\}$ is the ground truth affinity score for object pair $\{\mathbf{d}^t_k, \mathbf{o}^{t-1}_m\}$. To see our ground truth label generation process, please refer to Sec.~\ref{details}.

Note that the above tasks are inherently intertwined and simultaneously optimized via end-to-end training. Supervising both scene flow estimation and motion segmentation directly aids the computation of the affinity matrix, which utilizes scene flow and clusters of moving points as input. Conversely, the gradients originating from the affinity matrix loss instruct the backbone to encode potent features from point clouds. This indirect guidance subsequently enhances both motion segmentation and scene flow estimation.

%% file: section/experiment.tex
\section{Experiments}\label{exp}

\subsection{Evaluation Settings}\label{settings}

\noindent\textbf{Dataset.} In our experiments, we demonstrate the effectiveness of \sysname using the View-of-Delft (VoD) dataset~\cite{palffy2022multi}, which includes essential components (i.e., 4D radar point clouds, odometry information, object bounding boxes and tracking IDs annotations) for our problem. As an official benchmark specific for 3D object detection, the annotations of its test split are not publicly available, thereby we evaluate our trained models with its validation split, which is unseen during our training process.

\noindent\textbf{Evaluation metrics.} To quantify our performance, we use the classical MOTA, MODA, MT, ML metrics~\cite{leal2015motchallenge, bernardin2008evaluating} and the popular sAMOTA, AMOTA, AMOTP metrics~\cite{weng20203d} for evaluation. To make these metrics adapt to our cluster-based object detections, we compute the IoU by counting the number of intersected and united radar points between the ground truth object and the predicted one. The threshold for our point-based IoU is set as 0.25 across all experiments.

\noindent\textbf{Baselines.} As there are no prior works for 4D radar-based moving object tracking, we select two state-of-the-art LiDAR-oriented 3D MOT methods, i.e., AB3DMOT~\cite{weng20203d} and CenterPoint~\cite{yin2021center} as our baselines. To ensure the comparison is fair, we keep their original settings and also train their models on the VoD training split. Note that baseline methods, though designed for LiDAR, also take 4D radar point clouds as input in this work. Specifically, we develop two augmented baselines AB3DMOT-PP and CenterPoint-PP by replacing the detector with PointPillars~\cite{lang2019pointpillars} for AB3DMOT and the backbone with that of PointPillars for CenterPoint. 
% As our baselines detect and track 3D bounding boxes, we transform their object outputs into clusters of points so that their results can be compared with ours. 
%Specifically, we only use valid objects (containing five or more points) from both predictions and ground truth for evaluation. This contributes to a more equitable comparison between the two fashions of methods as the detectors in baselines are prone to miss objects with fewer points while our method does not. 
\begin{table*}[!htbp]\footnotesize
    \centering
    \caption{Performance of RaTrack and baselines on VoD. Baselines with (A) represent methods trained and evaluated on all objects, while others are trained and evaluated only on moving objects, which serve as the main comparison for RaTrack. } 
    \renewcommand\arraystretch{1}
    \setlength\tabcolsep{12pt}% 
    \resizebox{\textwidth}{!}{%
    \begin{tabular}{@{}lccccccc@{}}
    \toprule
    Method  & sAMOTA [\%] $\uparrow$ & AMOTA [\%] $\uparrow$ & AMOTP [\%] $\uparrow$ &  MOTA [\%] $\uparrow$ & MODA [\%] $\uparrow$ & MT [\%] $\uparrow$ & ML [\%] $\downarrow$ \\
    \midrule
    CenterPoint~\cite{yin2021center} (A) & 37.72 & 8.77 & 38.92 & 33.34 & 35.17 & 11.73 & 57.41   \\
    CenterPoint-PP~\cite{yin2021center, lang2019pointpillars} (A) & 42.64 & 10.87 & 43.64 & 36.20 & 37.29 & 15.43 & 50.62   \\
    AB3DMOT~\cite{weng20203d} (A) & 33.56 & 6.66 & 33.34 & 31.00 & 31.20 & 14.81 & 64.20  \\
    AB3DMOT-PP~\cite{weng20203d, lang2019pointpillars} (A) & 35.82 & 7.67 & 36.70 & 38.44 & 41.96 & 19.12 & 38.24  \\
    \midrule
    CenterPoint~\cite{yin2021center} & 43.21 & 14.40 & 54.55 & 38.44 & 41.96 & 19.12 & 38.24  \\
    CenterPoint-PP~\cite{yin2021center, lang2019pointpillars} & 44.54 & 16.33 & 58.80 & 43.96 & 44.91 & 19.12 & 54.41 \\
    AB3DMOT~\cite{weng20203d} & 51.23 & 15.00 & 53.21 & 46.72 & 47.38 & 20.59 & 39.71  \\
    AB3DMOT-PP~\cite{weng20203d, lang2019pointpillars} & 60.71 & 21.51 & \textbf{62.75} & 49.38 & 49.86 & 26.47 & 33.82  \\
    \midrule
    \textbf{RaTrack} & \textbf{74.16}  & \textbf{31.50}  & 60.17 & \textbf{67.27}  & \textbf{77.83} & \textbf{42.65} & \textbf{14.71}   \\
    \bottomrule
    \end{tabular}
    \label{main_table}
    }
    % $\dagger$ indicates that the sAMOTA metric adjusts its lower bound to zero for methods with negative MOTA values. } 
    \vspace{-1em}
\end{table*}

\begin{figure*}[!tbp]
  \centering
  \includegraphics[scale=0.15]{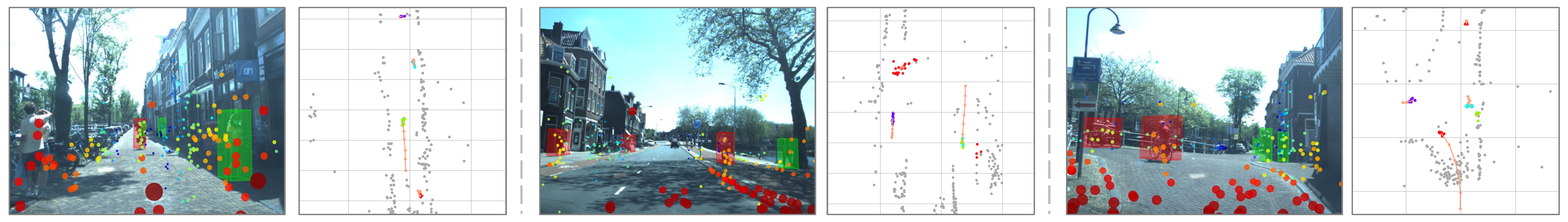}
  \vspace{-0.5em}
  \caption{Qualitative results of \sysname. We show on RGB images the ground truth 3D bounding boxes of moving objects (colors indicate different object classes) and projected radar point clouds (colors indicate the distances of points). On corresponding bird's eyes view figures, we show the detected moving objects and their predicted trajectories, where different colors are used to distinguish multiple object clusters.}
  \label{qual}
  \vspace{-1em}
\end{figure*}

\subsection{Implementation Details}\label{details}

\noindent\textbf{Label generation.} We follow~\cite{baur2021slim, ding2022self,ding2023hidden,jund2021scalable} to generate \emph{pseudo} scene flow labels using the ego-motion and object annotations. The ground truth motion segmentation mask can then be obtained by thresholding after compensating the ego-motion from the scene flow. 
% Please refer to our supplementary materials for more details.
To get the ground truth affinity matrix, we first match ground truth objects with our detected objects. A detected object that has a point-based IoU higher than 0.25 with any ground truth object will be assigned the same ID as the ground truth. Then, the ground truth affinity matrix can be constructed by assigning $\tilde{a}^t_{k,m}=1$ if the object pair $\{\mathbf{d}^t_k, \mathbf{o}^{t-1}_m\}$ has the same ID, and verse vice.

\noindent\textbf{Hyperparameters.} Our hyperparameters are determined both empirically and according to the testing results. The threshold $\zeta_{mov}$ (c.f. Sec.~\ref{detection}) is simply set as 0.5. For the DBSCAN algorithm, the neighbourhood radius is 1.5m while the minimum number of points in a cluster is set as 2. For loss weights, we set $\alpha_1, \alpha_2, \alpha_3$ in the overall loss (c.f. Eq.~\ref{loss:overall}) to be 0.5, 0.5 and 1.0 respectively, and set $\beta$ in the motion segmentation loss (c.f. Eq.~\ref{loss:seg}) to be 0.4. 

\noindent\textbf{Network training.} Training our network is non-trivial due to the dependencies of intermediate outputs across modules and frames. For example, the efficacy of our data association module depends on whether moving objects are correctly detected in the object detection module. To more effectively train our network, we separate the training into two stages: 1) We first train the backbone and the class predictor used for motion segmentation with Eq.~\ref{loss:seg} and keep other components frozen. The training keeps for 16 epochs with an initial learning rate of
0.001. This allows for fast learning of accurate moving object detection before data association. 2) We then train the whole network end-to-end for an additional 8 epochs with an initial learning rate of 0.0008. 
The Adam optimizer~\cite{kingma2014adam} is used for network parameter updates and the learning rate decays by 0.97 per epoch in both stages.

\subsection{Overall Performance} 

We evaluate \sysname and our baseline methods and compare their results on different metrics, as shown in Table~\ref{main_table}. \sysname exhibits a remarkable gain over baselines, increasing the sAMOTA, AMOTA and MOTA by 13.4\%, 10.0\%, 17.9\% compared to the second-best scores. Such results demonstrate its superiority in 4D radar-based moving object detection and tracking. Notably, \sysname achieves an improvement of 28.0\% on the MODA metric that is used to measure the object detection accuracy. This supports that our class-agnostic object detection without bounding boxes is a better option than current 3D bounding box detectors~\cite{yin2021center,lang2019pointpillars, shi2019pointrcnn} for recognizing and localizing moving objects in 4D radar point clouds. With more reliable object detection and data association module, \sysname also surpasses all baselines on the MT and ML metric, which demonstrates its ability to maintain long-term tracking of moving objects. It can be also observed that \sysname has a slightly lower AMOTP than the AB3DMOT-PP baseline~\cite{weng20203d,lang2019pointpillars}. As a precision metric, AMOTP only calculates the IoU between successfully matched object pairs, thus it becomes less important when the MOTA and MODA, which assess the tracking and detection accuracy, are considerably lower. We also show some qualitative results of \sysname in Fig.~\ref{qual}.

\begin{table*}[!htbp]\small
    \centering
    \caption{Ablation study results for RaTrack. For the first row, the bounding box detection results from PointPillars~\cite{lang2019pointpillars} are employed to generate object clusters as input to data association.}
    \renewcommand\arraystretch{1}
    \setlength\tabcolsep{12pt}% 
    \resizebox{\textwidth}{!}{%
    \begin{tabular}{@{}lcccccccc@{}}
    \toprule
     Method & sAMOTA [\%] $\uparrow$ & AMOTA [\%] $\uparrow$ & AMOTP [\%] $\uparrow$ &  MOTA [\%] $\uparrow$ & MODA [\%] $\uparrow$ & MT [\%] $\uparrow$ & ML [\%] $\downarrow$ \\
    \midrule
    Replace ODM with PP~\cite{lang2019pointpillars} & 53.12 & 17.47 & 53.65 & 41.66 & 42.32 & 24.21 & 38.95  \\
    Remove MEM & 68.45 & 26.08 & 51.23 & 62.32 & 71.27 & 38.24 & 16.18  \\
    Remove the velocity & 31.50 & 5.63 & 16.83 & 24.17 & 32.45 & 11.76 & 30.88 \\
    \midrule
       \textbf{RaTrack} & \textbf{74.16}  & \textbf{31.50}  & \textbf{60.17} & \textbf{67.27}  & \textbf{77.83} & \textbf{42.65} & \textbf{14.71}   \\
    \bottomrule
    \end{tabular}
    }
    \label{ablation}
    \vspace{-1em}
\end{table*}

As seen in Table~\ref{main_table}, we also train and evaluate baseline models with all annotated objects rather than only the moving ones to compare their performance in two settings. A consistent performance gap can be observed between the two settings of our baselines, which demonstrates that detecting and tracking moving objects is easier than that of general objects. Moreover, we find that our two augmented baselines (i.e., with -PP) outperform the original ones. We credit this to PointPillars~\cite{lang2019pointpillars} which is the state-of-the-art for 4D radar-based 3D object detection as exhibited in~\cite{palffy2022multi}.

\subsection{Ablation Study}
To validate the effectiveness of the object detection module (ODM) (c.f. Sec.~\ref{detection}) and motion estimation module (MEM) (c.f. Sec.~\ref{motion}), here we conduct an ablation study to see their impact. We also analyze the impact of the auxiliary velocity features. The results are shown in Table~\ref{ablation}. 

\noindent\textbf{Object detection module.} By replacing our ODM with PointPillars detector~\cite{lang2019pointpillars}, the performance on sAMOTA, MOTA and MODA degrades by 21.0\%, 25.6\% and 35.5\% respectively. This supports again that our cluster-based object detection is more effective than previous detectors for moving object detection and can thus yield more accurate tracking results. However, the outcome of this ablated version is less optimal than anticipated. We credit this to the incompatibility of the bounding box detector with our framework. Specifically, the object clusters derived from bounding boxes might have more noisy points, which further interferes with our data association that utilizes point-level features. 
% Second, without the supervision of motion segmentation, the feature learning in the backbone will be implicitly affected.  

\noindent\textbf{Motion estimation module.} The removal of our MEM yields the decrease of 5.7\%, 5.4\%, 6.6\% on sAMOTA, AMOTA and MODA. Such a change in performance demonstrates that our estimated scene flow from MEM can facilitate robust object detection and benefit object temporal matching. Specifically, without MEM, our AMOTP drops by 8.9\% which is a non-trivial change in the detection precision. This highlights the importance of our scene flow estimation as an additional motion cue in clustering. Points with similar scene flow vectors are prone to be grouped together.

\noindent\textbf{Auxiliary velocity feature.} By incorporating the velocity information into the input, a substantial gain is witnessed in all metrics, which demonstrates that the input velocity features are indispensable for \sysname. Indeed, the velocity features are the key enabler to scene flow estimation and motion segmentation tasks in our network by providing point-level motion information to be encoded in the backbone features. On the other hand, \sysname can sufficiently utilize such features with its bespoken network designed for 4D radar.

\begin{figure}[!tbp]
  \centering
  \includegraphics[scale=0.285]{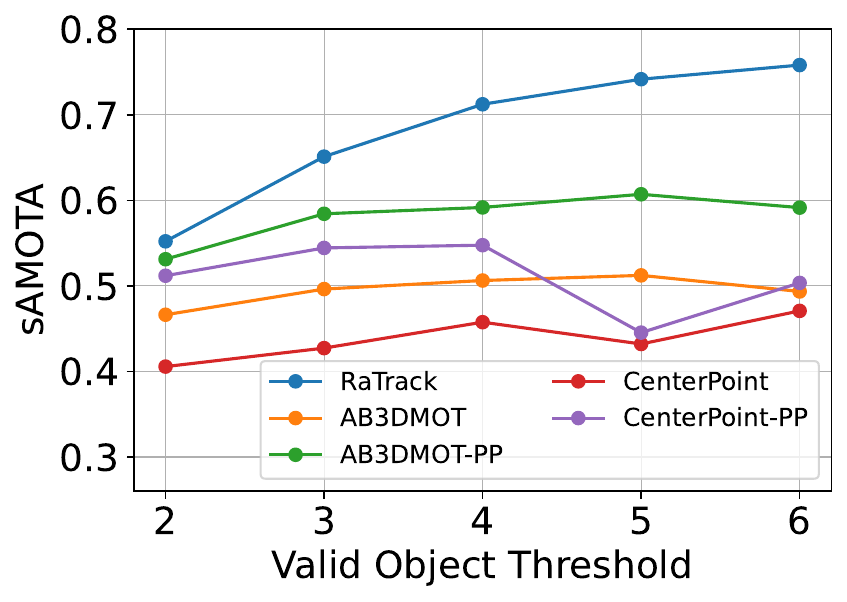}
  \includegraphics[scale=0.285]{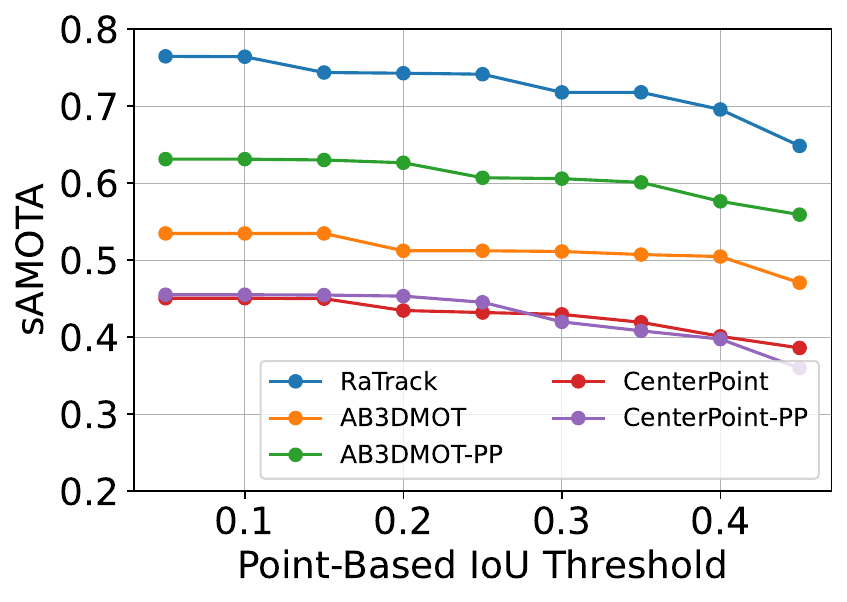}
  \caption{Result comparison between \sysname and baselines on varying valid object thresholds (both predicted and ground truth objects are filtered) and point-based IoU thresholds.}
  \label{threshold}
  \vspace{-1em}
\end{figure}

\subsection{Sensitivity Analysis}

\noindent\textbf{Impact of valid object threshold.} 
During the evaluation, we ignore invalid objects ($<5$ points) from both predictions and ground truth as we are more interested in the objects that are \emph{sufficiently} measured. Here we investigate the impact of this valid object threshold (i.e., the minimum number of points to identify an object as valid) on our evaluation results. As exhibited in Fig.~\ref{threshold} (left), \sysname consistently outperforms all baseline methods, regardless of the valid object threshold enlarges, and continuously increase its scores as the threshold. We credited this to our cluster-based object detection that can group any number of neighbour points into objects for tracking, where objects with more points are more easily being recognized and tracked. In contrast, it is hard for our baseline methods to produce comparable results with 3D bounding box-based detection strategy.

\noindent\textbf{Impact of point-based IoU threshold.} We also analyze the impact of the point-based IoU threshold (i.e., the minimum IoU to be identified as a true positive sample) for our evaluation results. As seen in Fig.~\ref{threshold} (right), \sysname achieves the best results on all IoU thresholds with a consistent improvement of $\sim 10\%$ over baselines, which further confirms the superiority of our pipeline. 

\begin{figure}[!tbp]
  \centering
  \includegraphics[scale=0.25]{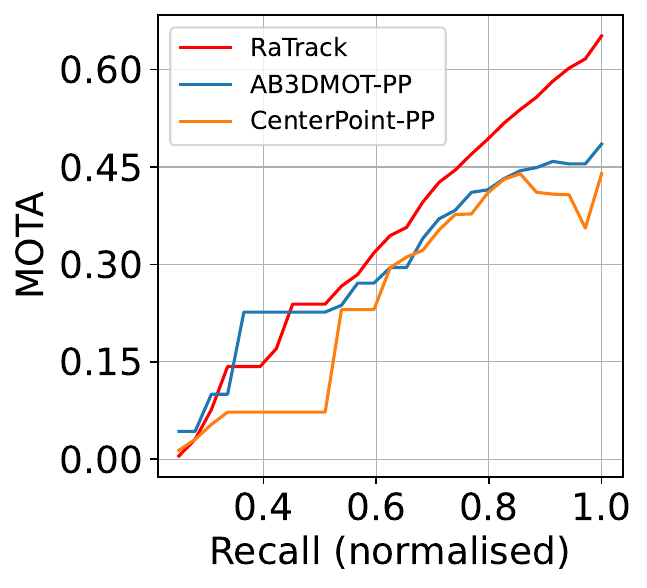}
  \includegraphics[scale=0.25]{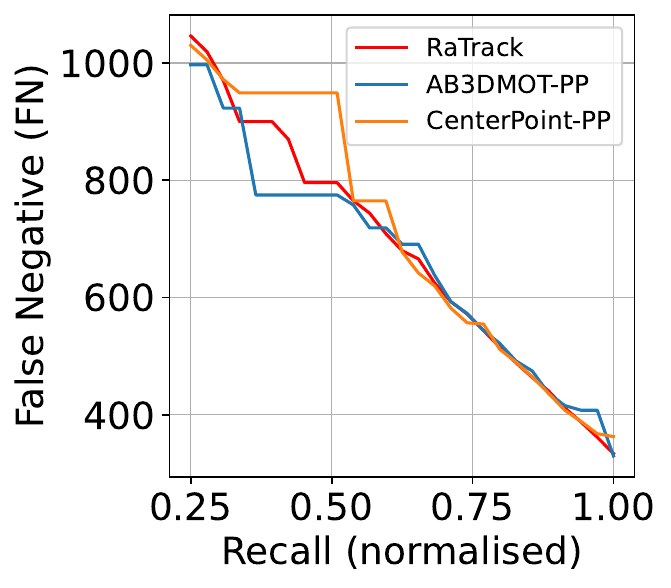}
  \includegraphics[scale=0.25]{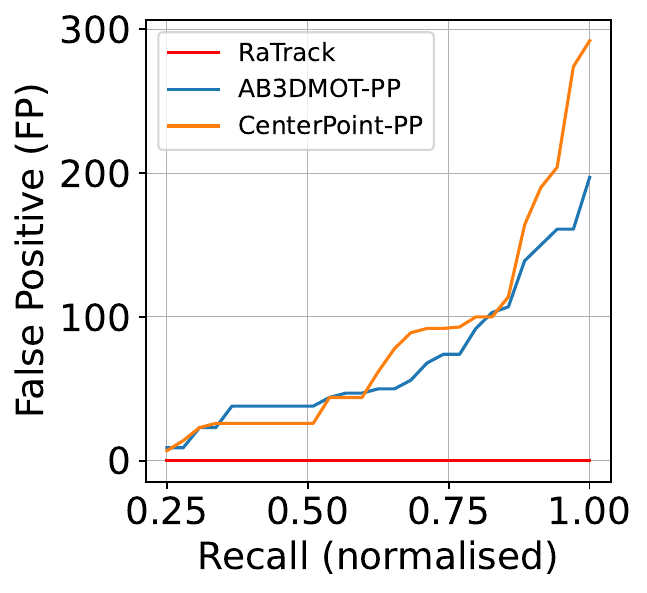}
  \caption{The effect of confidence values on the MOTA and the number of FN and FP. As each method has its own range of recall, we normalised their results from 0.25-1.0 for comparison. }
  \label{recall}
  \vspace{-1em}
\end{figure}

\noindent\textbf{Impact of confidence threshold.} When calculating our metrics, AMOTA and MOTA, the results need to be evaluated repetitively on different confidence thresholds, which further corresponds to different recall values~\cite{bernardin2008evaluating,weng20203d}. Then the best MOTA score is reported for comparison and the average MOTA is computed as the AMOTA score. Here we show the impact of the confidence threshold on the MOTA and the number of FN and FP, which are two key components to compute the MOTA. As seen in Fig.~\ref{recall}, \sysname shows higher MOTA scores over most recall values compared to two augmented baselines. Notably, after ignoring objects with less than five points in evaluation, \sysname hardly exhibits any FPs, avoiding the compromises in balancing FPs and FNs faced by the baseline methods.

%% file: section/conclusion.tex
\section{Conclusion}

In this work, we unveiled the untapped potential of 4D mmWave radars for multiple moving object tracking. Addressing the challenges of noise and point sparsity in radar data, our approach, \sysname, offers a fresh perspective on the tracking of moving objects, emphasizing the utility of motion segmentation and clustering over the conventional dependence on specific object types and bounding boxes. This restructured approach not only simplifies the process but also boosts the accuracy of multi-object tracking in complex dynamic scenarios. Extensive evaluations on a public dataset highlight the method's competitive performance.